\documentclass[journal]{IEEEtran}
\usepackage{amsmath,graphicx,epstopdf}
\usepackage{epsfig}
\usepackage{citesort}
\usepackage{multirow,bigstrut}
\usepackage{amssymb}
\usepackage{color}
\usepackage{url}
\usepackage{booktabs}
\usepackage{diagbox}
\usepackage{subfigure}

\begin{document}
\title{Deep Reference Generation with Multi-Domain Hierarchical Constraints for Inter Prediction}

\author{Jiaying~Liu, ~\IEEEmembership{Senior Member,~IEEE,}
        Sifeng~Xia,
        Wenhan~Yang, ~\IEEEmembership{Member,~IEEE}
}

 %\markboth{IEEE TRANSACTIONS ON MULTIMEDIA, DRAFT}%
% {Shell \MakeLowercase{\textit{et al.}}: Bare Demo of IEEEtran.cls for IEEE Journals}

\maketitle

\begin{abstract}
Inter prediction is an important module in video coding for temporal redundancy removal, where similar reference blocks are searched from previously coded frames and employed to predict the block to be coded.
Although traditional video codecs can estimate and compensate for block-level motions, their inter prediction performance is still heavily affected by the remaining inconsistent pixel-wise displacement caused by irregular rotation and deformation.
%the remaining inconsistent pixel-wise displacement caused by irregular rotation and deformation still heavily affect the inter prediction performance.
In this paper, we address the problem by proposing a deep frame interpolation network to generate additional reference frames in coding scenarios.
First, we summarize the previous adaptive convolutions used for frame interpolation and propose a factorized kernel convolutional network to improve the modeling capacity and simultaneously keep its compact form.
Second, to better train this network, multi-domain hierarchical constraints are introduced to regularize the training of our factorized kernel convolutional network.
For spatial domain, we use a gradually down-sampled and up-sampled auto-encoder to generate the factorized kernels for frame interpolation at different scales.
For quality domain, considering the inconsistent quality of the input frames, the factorized kernel convolution is modulated with quality-related features to learn to exploit more information from high quality frames.
For frequency domain, a sum of absolute transformed difference loss that performs frequency transformation is utilized to facilitate network optimization from the view of coding performance. With the well-designed frame interpolation network regularized by multi-domain hierarchical constraints,
our method surpasses HEVC on average $6.1\%$ BD-rate saving and up to $11.0\%$ BD-rate saving for the luma component under the random access configuration.
\end{abstract}

\begin{IEEEkeywords}
High Efficient Video Coding (HEVC), inter prediction, frame interpolation, deep learning, multi-domain hierarchical constraints, factorized kernel convolution
\end{IEEEkeywords}

\IEEEpeerreviewmaketitle

\section{Introduction}
\label{intro}
%\IEEEPARstart{T}{here} are high temporal similarities among video frames.

\IEEEPARstart{W}{ITH} the booming multimedia social networking and consumer electronics markets, a tremendously increasing amount of images and videos are uploaded to the community everyday. The new trend calls for new coding techniques to further improve the compression efficiency. Successive video frames are usually continuous in the temporal dimension and capture the same scene.
Therefore, video codecs like MPEG-4 AVC/H.264 \cite{Overviewavc} and High Efficiency Video Coding (HEVC) \cite{Overviewhevc} seek to improve the video coding performance with inter prediction by removing temporal redundancy between video frames.
Specifically, in the inter prediction module, for a block which is to be coded (to-be-coded block), the motion estimation technique is first used to search for reference blocks among the reconstructed frames. Based on the motion estimation results, motion compensation technique then predicts the to-be-coded block based on reference blocks. After that, only the block-level motion information and the prediction residue between the predicted result and the original to-be-coded-block need to be coded. Consequently, temporal redundancies are largely removed and many bits can be saved.

However, there are lots of obstacles to performing the inter prediction. Even for continuous frames, content changes and complex local motions are quite common, which lead to large residues. Thus, many bits are used to code these residues between the prediction and the to-be-coded block.
Many researches are conducted to better estimate global and local motions, namely capturing inter-frame correspondences, for better motion compensation and temporal redundancies removal.
The early works~\cite{Overviewavc,Overviewhevc} start to perform \textit{block-level motion estimation and compensation}. In these methods, the prediction is derived directly from one individual reference block or a linear combination of the reference blocks.
In real videos, besides block-level translational motion between reference blocks and the to-be-coded block, there exist complex local motions caused by non-translational camera and object movements, which are called \textit{inconsistent pixel-wise displacement}, like rotation and deformation between the matched blocks.
These kinds of inconsistent pixel-wise displacement cannot be modeled only by block-level motion estimation and compensation.
The residues are still large and cost a lot of bits for coding.

Therefore, some methods \cite{bof2010,bof2016,cefi} turn to applying \textit{pixel-wise refinement} for inter prediction with the bi-directional optical flow (BIO) between reference frames. Alshin \textit{et al.} \cite{bof2010,bof2016} calculated BIO between reference blocks and operated pixel-wise motion refinement for the bi-directional motion compensation. In \cite{cefi}, with the BIO estimated from reference frames, a co-located reference frame is interpolated as the additional reference for motion compensation. Although alleviating pixel-wise displacement to some extent, the performance of these methods heavily relies on the accuracy of optical flow estimation. However, the optical flow estimation process used by these methods is manually designed, which will inevitably lead to the inaccurate estimation of the complex pixel-wise inconsistent displacement.
% owing to the time efficiency requirement in video coding tasks, the optical flow estimation accuracy of the above methods is usually not satisfying.

Recently, with the rise and development of deep learning-based image processing, some researchers begin to devote their efforts to utilizing deep learning techniques to address motion-related problems, \textit{e.g.} optical flow estimation \cite{flownet2,spynet,PWCNet} and frame interpolation \cite{phasenet,Niklaus_CVPR_2017,Niklaus_ICCV_2017}. Besides significantly improving the performances in these tasks, these works bring in new insights and methodologies for pixel-level motion modeling, which provide new foundations for the successive works. Meanwhile, more and more works \cite{kimicip17,loopjia,liTIP18,halfpel} explore to introduce deep learning techniques to the video coding scenario and offer significant improvements in coding performance. Due to the powerful capacity of representation learning, deep learning techniques can flexibly handle various kinds of video signals and successfully construct the non-linear mapping from input signals to the target domain.

We follow both trends, deep learning-based motion modeling and deep learning-based video coding optimization, and offer optimized video coding techniques to better model the pixel-wise inconsistent displacement.
Specifically, we choose to use deep learning techniques to interpolate a pixel-wise closer frame (PC-frame) from existing reconstructed frames. Here, ``pixel-wise closer'' means that the inconsistent pixel-wise displacement between the interpolated frame and the the frame which is to be coded (to-be-coded-frame) is smaller than that between the reconstructed frames and the to-be-coded frame. After that, the PC-frame is utilized as an additional reference frame for the to-be-coded frame. Thereby reference blocks with smaller pixel-wise displacement may be retrieved for the to-be-coded blocks in inter prediction.
Compared to the individual video frame interpolation task, frame interpolation in video coding additionally faces more issues.
1) In the lossy compression, reconstructed reference frames are heavily degraded so less reference information could be used for interpolation. Moreover, the interpolation of the detail will be disturbed by compression artifacts in the prediction process. 2) Coding performance should be considered as the metric for coding-oriented frame interpolation.
However, the existing coding pipeline is very complex and not end-to-end trainable. So it is also a great challenge to introduce proper objective functions to train the coding-oriented frame interpolation network.
3) There are various kinds of dependencies in different domains which can be utilized for PC-frame interpolation, \textit{e.g.} spatial domain, frequency domain, \textit{et al.} There is not a unified framework to consider these dependencies and their potential interactions jointly.

In our work, we tackle the above issues by building a multi-scale quality attentive factorized kernel convolutional neural network (MSQ-FKCNN).
The network exploits an encoder-decoder convolutional neural network (CNN) to generate factorized kernels for synthesizing the target frame from compressed frames.
Compared with a single large kernel or separable kernels, the proposed network is both flexible and economic to model video frame signals with factorized kernels.
Meanwhile, we introduce multi-domain hierarchical constraints to train the network. 1) To reduce the disturbance of compression noise, we introduce a quality attentive mechanism which guides the network to make choices in the \textit{quality domain} to use more information from high quality frames for inter prediction. 2) For the metric to train such a network, inspired by HEVC, a sum of absolute  transformed difference (SATD) loss function that integrates measurements in both \textit{spatial} and \textit{frequency domains} is used. 3) To better utilize dependencies in the \textit{spatial domain} and model \textit{the joint interdependencies among different domains}, our network takes a multi-scale structure to exploit the spatial dependencies and model the multi-domain dependencies in a unified way.
Benefiting from our well-designed factorized kernel CNN and the carefully considered multi-domain hierarchical constraints, the proposed network can be trained not only for better interpolation quality but also greater coding performance.

% This paper is an extension of our previous conference paper \cite{PMODFI}.
% \M{Based on the preliminary work, we additionally make the following endeavors.
% First, we intensively analyze challenges faced in the coding-oriented frame interpolation. Second, based on the analysis, to tackle the faced issues, we propose an improved framework to predict a PC-frame under the hierarchical multi-domain constraints.
% The multi-scale quality attentive factorized kernel CNN is formulated and employed to implement the desired coding-oriented frame interpolation.
% A quality attentive mechanism is additionally introduced to make the network pay more attention to input frames of higher qualities to alleviate the disturbance of compression artifacts.
% We further conduct more extensive experimental analysis to identify the effectiveness of our method and its each component. In summary,}

Our contributions are summarized as follows:

\begin{itemize}
\item We propose to utilize deep frame interpolation to generate an additional pixel-wise closer reference frame for inter prediction. A coding-oriented frame interpolation network MSQ-FKCNN is specially designed to flexibly synthesize the target frame from input frames with factorized kernels.
MSQ-FKCNN significantly alleviates the inconsistent pixel-wise displacement between existing reference frames and the to-be-coded frame.

\item To better train our network, multi-domain hierarchical constraints are designed for the coding-oriented frame interpolation. The hierarchical dependencies in spatial, quality and frequency domains are considered jointly to obtain more abundant reference information and achieve better interpolation results.

\item To additionally deal with compression artifacts, the multi-scale quality attentive mechanism is designed to make the network pick up more information from high quality frames and further exploit spatial dependencies for prediction, which further improves the interpolation accuracy.

\item In order to improve the modeling capacity of our network in the video coding scenario, a multi-scale SATD loss function is implemented to guide the network optimization in the joint spatial and frequency domain, which can better indicate the coding cost of the prediction residue and lead to better coding performance.

\end{itemize}

The rest of the paper is organized as follows. Sec. \ref{related works} introduces recently proposed deep learning-based methods which solve motion-related problems. Some recent works that use deep learning techniques to improve video coding performance are also presented. Our proposed coding-oriented frame interpolation method will be introduced in Sec. \ref{sec3}. Implementation details about the training data preparation and how to integrate generated PC-frames into HEVC are shared in Sec. \ref{sec4}. Experimental results and analyses are shown in Sec. \ref{sec5} and concluding remarks are given in Sec. \ref{sec6}.

\section{Related Works}
\label{related works}
% \subsection{Traditional Pixel-Wise Motion Oriented Inter Coding Methods}
\subsection{Deep Learning-Based Motion-Related Works}
Recently, deep learning-based motion estimation works have been widely proposed and show impressive results compared with traditional methods. In \cite{flownet}, an end-to-end optical flow estimation network FlowNet is first proposed and achieves comparable estimation accuracy with traditional methods. A succeeding network FlowNet 2.0 is later designed to progressively estimate the optical flow and perform on par with state-of-the-art methods at higher frame rates. Recently, Sun \textit{et al.} \cite{PWCNet,PWCNetpami} proposed a compact but effective PWC-Net integrating pyramid processing, warping, and the cost volume. The proposed PWC-Net successfully outperforms previous works on the KITTI benchmark \cite{kitti}.

Meanwhile, some motion-related applications like frame interpolation are also greatly facilitated by deep learning techniques.
% researchers have achieved better frame interpolation performance with deep learning techniques.
Niklaus \textit{et al.} \cite{Niklaus_CVPR_2017} formulated video frame interpolation as two steps, \textit{i.e.} motion estimation and pixel synthesis, and they proposed an end-to-end deep learning framework to solve these two tasks. Spatially adaptive kernels are estimated for synthesizing target frames. In \cite{Niklaus_ICCV_2017}, adaptive separable kernels are successively proposed to largely reduce the model parameters. Liu \textit{et al.} \cite{voxelflow} choosed to directly synthesize the target frame from the input by learning pixel displacement with the network. In \cite{CSVFI}, flows between the target frame and two input frames are also estimated and utilized for warping. The warped contextual information which is extracted from the response of ResNet-18 \cite{resnet} is additionally used for blending intermediate frames warped from two-sided input frames. In \cite{superslomo},  bi-directional optical flows between input frames are inferred by U-Net \cite{unet} and then linearly combined at each time step for interpolating target frames at arbitrary time points.

The successes of all the above deep learning-based methods have identified the ability of deep learning techniques in handling motion related problems. Thus, based on the meaningful experiences of previous methods, we make deeper explorations to estimate the pixel-wise displacement between compressed video frames and generate better temporal reference samples for inter prediction in video coding with deep neural networks.

\subsection{Deep Learning-Based Video Coding}

There has been a bunch of works exploiting deep learning techniques to improve video coding performance by optimizing modules in the coding structure, \textit{e.g.}, loop filtering, mode decision, rate control, intra and inter prediction.

CNN has brought significant performance gain to many image restoration tasks like super-resolution \cite{SRCNN,VDSR}, denoising \cite{dncnn} and compression artifacts removal \cite{arcnn,dai2017convolutional}. The success of CNN on these image restoration tasks has promoted the development of deep learning-based loop filtering methods.
In \cite{kimicip17}, Kang \textit{et al.} proposed a multi-modal/multi-scale convolutional neural network to replace existing deblocking filter and sample adaptive offset for loop filtering, which obtains considerable gains over HEVC. A content-aware mechanism \cite{jiaTIP19} is designed to use different CNN models for the adaptive loop filtering in different regions. In addition to improving loop filtering performance, Laude and Ostermann \cite{laude2016deep} proposed to replace the conventional Rate Distortion Optimization (RDO) with CNN for the intra prediction mode decision. Furthermore, coding unit (CU) partition mode decision can also be predicted by CNN \cite{liu2016cu} and Long and Short-Term Memory (LSTM) network \cite{xu2017reducing}.

Considering the strong nonlinear mapping ability of deep learning techniques, it is also very promising to predict more reference signals from existing reconstructed signals for intra and inter prediction by deep learning. Li \textit{et al.} \cite{liTIP18} firstly adopted fully connected network (FCN) to learn an end-to-end mapping from neighboring reconstructed pixels to the to-be-coded bolck in the intra coding of HEVC. Moreover, Hu \textit{et al.} \cite{huvcip18} used a recurrent neural network to explore the correlations between reconstructed reference pixels and predict the to-be-coded block in a progressive manner. As for inter prediction, Wang \textit{et al.} \cite{wangicme18} additionally used spatially neighboring pixels of both reference blocks and current to-be-coded blocks to refine initial predicted blocks with an FCN and a CNN. In \cite{halfpel} and \cite{oneforall}, CNNs are used for fractional interpolation in the motion compensation process, which provide better sub-pixel level reference samples for inter prediction. Zhao \textit{et al.} \cite{fruc} first tried to apply deep frame interpolation to video coding by directly using interpolated blocks as the reconstructed blocks at coding tree unit (CTU) level. They directly applied a pre-trained video frame interpolation model without any specific optimization for the video coding scenario. Comparatively, in our work, we exploit multi-domain hierarchical constraints for the additional reference generation to overcome new challenges faced in the video coding scenario.

\section{Pixel-Wise Closer Reference Generation with Hierarchical Constraints in Multiple Domains}
\label{sec3}

In this section, we first illustrate the frame interpolation in HEVC and analyze several issues faced in the coding scenario.
Then, we build an MSQ-FKCNN for deep frame interpolation. At last, we present several well-designed constraints to regularize the training of our MSQ-FKCNN to address the above issues for better interpolation.

\subsection{Frame Interpolation in HEVC}
\label{sec31}

We implement and test our method on the HEVC reference software HM-16.15 under the RA configuration. For a to-be-coded frame $I_t$, two-sided frames $I_l$ and $I_r$ are previously coded and used as the input of MSQ-FKCNN. The PC-frame $I_m$ will be interpolated to facilitate inter prediction.
In the video coding scenario, only reconstructed reference frames ${\hat I}_l$ and ${\hat I}_r$ are available for reference. Compared to frame interpolation of high-quality videos, frame interpolation in the coding scenario inevitably faces four issues:
\begin{enumerate}
	\item \textit{High frequency detail loss} of reference frames caused by the quantization operation leads to difficulty in the implicit motion estimation and inaccuracy of local details inference.
	\item \textit{Compression artifacts}, \textit{i.e.} the blockiness, in reference frames originate from block-based quantization. The artifacts are easy to be brought into the generated interpolation results.
	\item \textit{Inconsistent quality of input frames}.
	Due to the design of coding configurations, $I_l$ and $I_r$ may be coded with different QPs. A desirable frame interpolation model should consider the quality of the input frames and utilize this information adaptively.
	\item \textit{The purpose of video coding} is to maintain the quality of decoded frames with fewer bits.
	Thus, training a coding oriented frame interpolation model should pay attention to both distortion and bit cost.
\end{enumerate}

\subsection{Overview of the Proposed Method}

To tackle the above mentioned issues, we explore possible models and potential constraints to effectively infer temporally intermediate frames from noisy and inconsistent input frames. To build an effective interpolation model, we start from raw adaptive kernel CNN and separate kernel CNN~\cite{Niklaus_ICCV_2017}, analyze their correlation with a unified viewpoint, and develop the proposed MSQ-FKCNN for better interpolation. To better train this model, the hierarchical constraints in several domains are introduced:
\begin{enumerate}
	\item \textit{Spatial Domain}.
	Our feature extraction network takes an encoder-decoder structure that first down-samples features and then up-samples features. In this network, the kernels and interpolation results are inferred from small to large progressively.
	At a small scale, the motion information is easier to be learnt and details can be better inferred in this progressive way even with the high frequency detail loss.
	Furthermore, at the small scale, compression artifacts are suppressed, and more clean and accurate interpolation results are obtained.
	\item \textit{Quality Domain}.
	To handle the inconsistent quality of input frames, we make our model be aware of the quality differences. The factorized kernel is modulated with quality-related features, which guides the model to utilize more information of the high quality input frame.
	\item \textit{Frequency Domain}. To better regularize the training of our frame interpolation network for better video coding performance, a loss considering both distortion as well as the bit cost is implemented by frequency transformation.
\end{enumerate}
In the following sections, we will introduce our MSQ-FKCNN and the multi-domain hierarchical constraints in details.

\begin{figure}
	\centering
	\includegraphics[width=8.5cm]{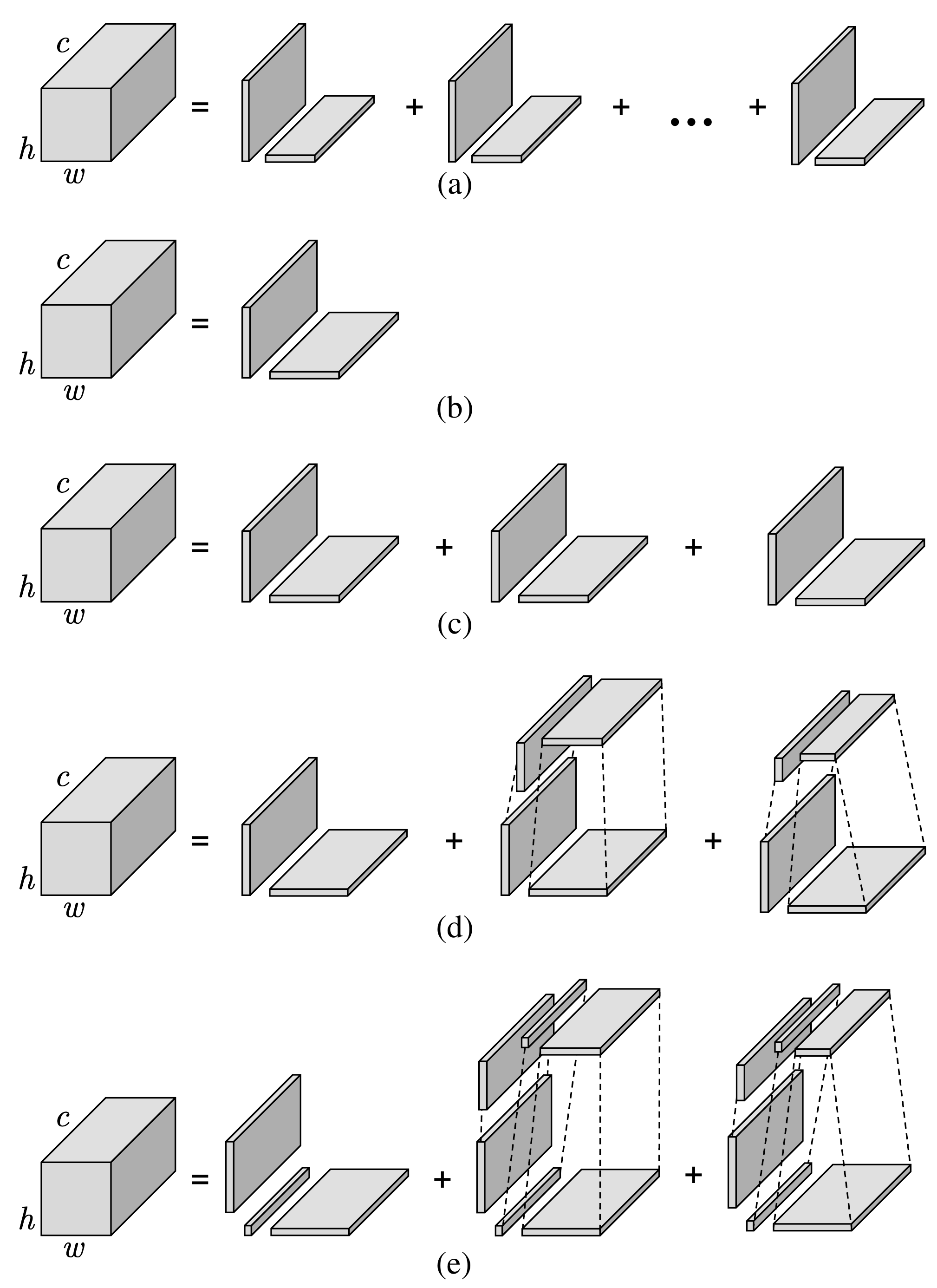}
	\caption{Architecture of different adaptive kernel models for frame interpolation.
	(a) Raw adaptive convolution and its factorized norm.
	(b) Adaptive separable convolution.
	(c) Factorized convolution.
	(d) Multi-scale factorized convolution.
	(e) Multi-scale quality attentive factorized convolution.
	}
	\label{fig:FK}

\end{figure}

\begin{figure*}[t]
	\centering
	\includegraphics[width=18cm]{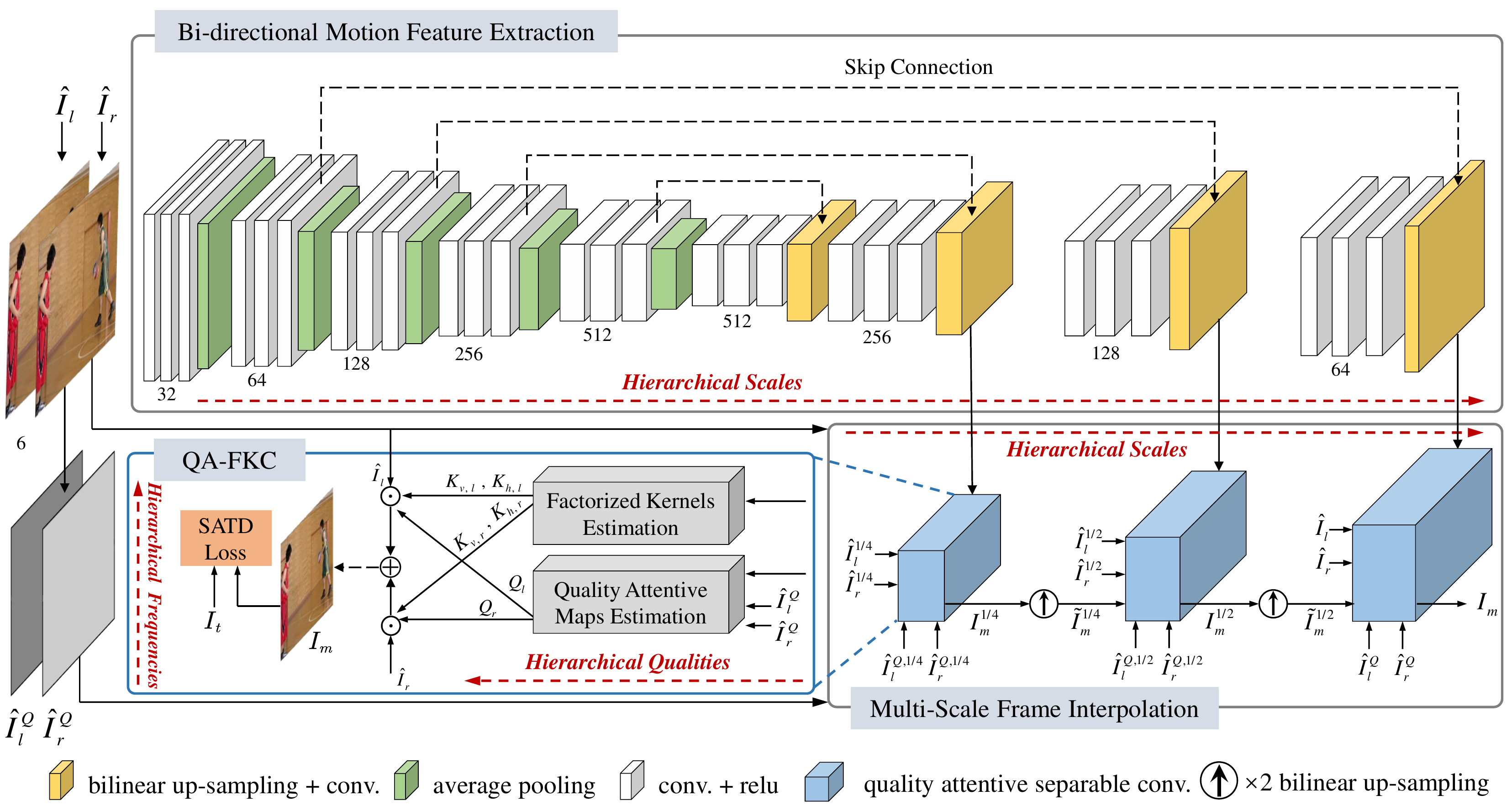}
\vspace{-2mm}
	\caption{
	Architecture of MSQ-FKCNN. Numbers below the feature maps indicate channel numbers. $1/2$ and $1/4$ mean scales of the images.
	The feature extraction part predicts bi-directional motion information between input frames.
	In the multi-scale frame interpolation component, intermediate frames of different scales are interpolated with the estimated factorized convolution kernels and quality attentive maps.
	QA-FKC denotes quality attentive factorized kernel convolution.
	The convolution is modulated with quality-related features to be aware of using more information from high quality frames.
	SATD loss measures the difference in both spatial and frequency domains. The whole network is regularized by the constraints in the spatial, frequency and quality domains.
	}
	\label{fig:f2}
	\vspace{-4mm}
\end{figure*}

\subsection{Multi-Scale Quality Attentive Factorized Kernel CNN}
\label{QASepconv}

For a to-be-coded frame $I_t$, the frame interpolation method based on adaptive convolutions uses the two-sided reference frames ${\hat I}_l$ and ${\hat I}_r$ as input. The bi-directional motion feature is first extracted and then used for inferring adaptive kernels to reconstruct the temporally intermediate frame.
The adaptive convolutions used in previous works, the related variants and our newly proposed one are discussed as follows.

\noindent \textbf{Adaptive Convolution}. Adaptive convolution works in this way.
To predict a pixel ${I_m}(x,y)$ in the target frame,
two $n \times n$ adaptive 2D kernels ${w_l}(x,y)$ and ${w_r}(x,y)$ will be first estimated respectively for two-sided input reference frames ${\hat I}_l$ and ${\hat I}_r$.
${I_m}(x,y)$ is then interpolated via local adaptive convolution on ${\hat I}_l$ and ${\hat I}_r$ as follows:
\begin{equation}\label{eq:kernel}
{I_m}\left( {x,y} \right) = {w_l}\left( {x,y} \right)*{\hat p_l}\left( {x,y} \right) + {w_r}\left( {x,y} \right)*{\hat p_r}\left( {x,y} \right),
\end{equation}
where ${\hat p_l}\left( {x,y}\right)$ and ${\hat p_r}\left( {x,y}\right)$ are $n \times n$ patches in ${\hat I_l}$ and ${\hat I_r}$ centered at the position $(x,y)$. For better illustration in the following parts, we first introduce a factorized form of the adaptive convolution as shown in Fig.~\ref{fig:FK}~(a). We concatenate the 2D kernels ${w_l}(x,y)$ and ${w_r}(x,y)$ together to form a 3D adaptive kernel $W\left( {x,y} \right)$ for each pixel ${I_m}(x,y)$. We assume the size of $W\left( {x,y} \right)$ to be $ c \times h \times w  $, where $h$ and $w$ represent heights and weights of the adaptive kernel and are set equally to $n$. $c$ belongs to the temporal dimension and corresponds to the number of input reference frames. Then, the kernel can be respectively factorized along the temporal dimension as follows:
\begin{align}\label{eq:factorized_kernel}
W(x,y,\theta ) = \sum\limits_{i = 1}^\lambda  {K_v^i{{\left( {x,y,\theta } \right)}^{'}}*K_h^i\left( {x,y,\theta } \right)} ,
\end{align}
where $\theta$ indicates the sequence number of the input reference frame, and $\lambda$ is the rank number of the adaptive kernel. $K_v^i{{\left( {x,y} \right)}}$ and $K_h^i\left( {x,y} \right)$ are factorized kernels of size $c \times 1 \times n$.

\noindent \textbf{Adaptive Separable Convolution}.
The raw adaptive convolution with a large kernel size leads to a huge amount of parameters, which makes the model training less promising. The adaptive separable convolution~\cite{Niklaus_ICCV_2017} addresses the problem by estimating the separable form of the convolutions. In fact, the adaptive separable convolution can be viewed as the special case of the factorized form of the adaptive convolution when $\lambda=1$, as shown in Fig.~\ref{fig:FK}~(b). For each pixel ${I_m}(x,y)$ in the target frame, four $1 \times n$ one-dimensional kernels ${K_v}\left( {x,y,1} \right)$, ${K_h}\left( {x,y,1} \right)$, ${K_v}\left( {x,y,2} \right)$ and ${K_h}\left( {x,y,2} \right)$ will be first estimated. Then, two $n \times n$ adaptive kernels $W(x,y,1 )$ and $W(x,y,2 )$ are obtained by $W(x,y,1 ) = {K_v}\left( {x,y,1} \right)^{'}*{K_h}\left( {x,y,1} \right)$ and $W(x,y,2 ) = {K_v}\left( {x,y,2} \right)^{'}*{K_h}\left( {x,y,2} \right)$.
Promising frame interpolation results can be achieved by estimating the adaptive separable kernels.

\noindent \textbf{Factorized Kernel Convolution}.
When we relax the approximate rank number $\lambda$ and set $\lambda$ to an intermediate value, we can get the convolutions with different number of model parameters and modeling capacities, as shown in Fig.~\ref{fig:FK}~(c) with $\lambda=3$.

\noindent \textbf{Multi-Scale Factorized Kernel Convolution}.
We assume that coding artifacts can be alleviated by the down-scaling operation. Thus, more accurate synthesis results can be obtained at small scales. By constraining the interpolation process at small scales, the main structure of the target frame is better learned and the frame interpolation quality can be further improved.

In conjunction with multi-scale frame interpolation, we project the factorized kernels to different scales and build a multi-scale factorized kernel convolution as shown in Fig.~\ref{fig:FK}~(d) as follows:
\begin{align}
\label{eq:factorized_kernel_ms}
\scriptsize
W(x,y,\theta) = \sum\limits_{\left\{ {s = 1,\frac{1}{2},\frac{1}{4}} \right\}} {K_v^s{{\left( {x,y, \theta} \right)}^{'}}*K_h^s\left( {x,y, \theta} \right)},
\end{align}
where $s$ represents the scale of the factorized kernel. Here, the representation of the adaptive kernel $W(x,y)$ is not realized by kernel-wise summation and is equivalently injected into the frame generation process. That is, the separate kernels are directly used to interpolate the target frame successively at different scales and are combined by the fusion of the synthesized frames of different scales.

Specifically, for each scale, the target pixel is synthesized by:
\vspace{-3mm}
\begin{equation}
\footnotesize
\label{eq3c}
I_{\rm{m}}^s\left( {x,y} \right) = \sum\limits_{\theta = 1}^c {K_v^s{{\left( {x,y,\theta} \right)}^{'}}*K_h^s\left( {x,y,\theta} \right)
* {{\hat P}^s}\left( {x,y,\theta} \right)}   + \tilde I_{\rm{m}}^{s/2}\left( {x,y} \right),
\end{equation}
where $I_{\rm{m}}^s$ is the target frame and $s$ represents the corresponding scale of $1/4$, $1/2$ or $1$. ${\hat P}^s\left( {x,y,\theta} \right)$ is the reference patch centered at the position $(x,y)$.  $\tilde I_{\rm{m}}^{s/2}\left( {x,y} \right)$ is obtained by doubly up-sampling the previously interpolated $ I_{\rm{m}}^{s/2}$ and it will be set to 0 for $s=1/4$.

\noindent \textbf{Multi-Scale Quality Attentive Factorized Kernel Convolution}.
As mentioned above, under the RA configuration, two-sided reference frames will be of different quality since they are coded with different QPs. It is meaningful to pay more attention to the reference frame of higher quality. Consequently, the quality attentive mechanism is introduced to the factorized kernel convolution and a new quality attentive kernel ${Q^s}\left( {x,y} \right)$ of size $c \times 1 \times 1$ is added. The quality attentive modulation as shown in Fig.~\ref{fig:FK} (e) is formulated as follows,
\begin{equation}\label{eq:factorized_kernel_ms_qa}
\footnotesize
W(x,y, \theta) \! = \! \sum\limits_{\left\{ {s = 1,\frac{1}{2},\frac{1}{4}} \right\}} {{Q^s}\left( {x,y, \theta} \right)*(K_v^s\left( {x,y, \theta} \right)^{'}*K_h^s\left( {x,y, \theta} \right))}.
\end{equation}

From the view of frame interpolation, an illustration of the target frame synthesis that uses quality attentive factorized kernel convolution is shown as the QA-FKC component in Fig. \ref{fig:f2}. We generate normalized quantization parameter (QP) maps ${\hat I}_l^Q$ and ${\hat I}_r^Q$ of the two-sided reference frames as the additional input to make our network more aware of the quality differences between reference frames. Two-sided quality attentive kernels and factorized kernels are estimated for synthesizing the target frame.

The target pixel $I_{\rm{m}}^s\left( {x,y} \right)$ is obtained by:
\begin{align}
\label{eq3}
\scriptsize
I_{\rm{m}}^s\left({x,y} \right) = & \sum\limits_{\theta = 1}^c {Q^s}\left( {x,y,\theta} \right)*(K_v^s{{\left( {x,y,\theta} \right)}^{'}}*K_h^s\left( {x,y,\theta} \right)) \nonumber \\
&  *{{\hat P}^s} \left( {x,y,\theta} \right) + \tilde I_{\rm{m}}^{s/2} \left( {x,y} \right).
\end{align}

\subsection{Architecture of MSQ-FKCNN}
The architecture of MSQ-FKCNN is shown in Fig. \ref{fig:f2}. The whole pipeline is illustrated in details as follows.

\noindent \textbf{Bi-Directional Motion Feature Extraction}. An encoder-decoder structure is employed to extract bidirectional motion feature. The progressive down-sampling and up-sampling operations effectively enlarge the receptive fields so large scale motion can also be caught by MSQ-FKCNN. Kernel sizes of all convolutional layers are set to $3 \times 3$ and the rectified linear unit (ReLU) is utilized as the activation function. At the encoder side, average pooling is used for down-sampling. Bilinear interpolation is used for up-sampling at the decoder side. Skip connections are used here to bypass low-level information from the encoder side to the decoder side.

\noindent \textbf{Multi-Scale Frame Interpolation}. With the extracted bi-directional motion feature, the multi-scale frame interpolation part generates target intermediate frames of different scales from small to large at the decoder side. At each scale $s$, the target intermediate frame is interpolated by quality attentive factorized kernel convolution.

\noindent \textbf{ Quality  Attentive  Factorized Kernel Convolution}. Details of factorized kernels estimation have been described in  Sec. \ref{QASepconv}. At each scale $s$, two-sided factorized kernels $K_{v,l}^s$, $K_{h,l}^s$ $K_{v,r}^s$, $K_{h,r}^s$ and quality attentive maps $Q_l^s$, $Q_r^s$ are estimated for interpolation. Feature maps of corresponding scales extracted in the feature extraction part are used as input. For a target frame of size $H \times W$, four factorized kernel maps of size $H \times W \times n$ will be inferred by four layers of convolutions. For scales of $1/4$, $1/2$ and $1$, $n$ is respectively set to $13$, $25$ and $51$. Thereby, each pixel in the target frame $I_m^s$ can find four corresponding $1 \times n$ factorized kernels at the same position of four factorized kernel maps.

As for the quality attentive maps estimation, normalized QP maps of reference frames are generated and used as the input together with the extracted bi-directional motion feature. The normalized QP maps are derived by dividing QPs of reference frames with the value 51. Then, an $H \times W \times 2$ quality attentive map is estimated by four layers of convolutions. The interpolated result $I_m^s$ is obtained by quality attentive factorized kernel convolution on the reference frames as illustrated in Eq.~\eqref{eq3}.

After multi-scale frame interpolation, the multi-scale SATD loss function is used to measure the prediction error and guide optimization of network parameters, which is illustrated in Sec. \ref{secMSATD}. The corresponding components in our network which implement the multi-domain hierarchical constraints are summarized in Table~\ref{tab:hierarchical_constraints}.

\subsection{Multi-Scale SATD Loss Function}
\label{secMSATD}
In the training process, parameters of the network are optimized by back-propagating the gradient of the loss calculated between the interpolated frame $I_m$ and ground truth $I_t$. In deep video frame interpolation methods, the $\ell_1$ loss function is commonly adopted \cite{voxelflow,Niklaus_ICCV_2017,CSVFI,superslomo} to train the model for the order of better objective performance:
\begin{equation}\label{eq31}
{\ell_1}\left( {I_m,I_{t}} \right) = {\left\| {I_m - I_{t}} \right\|_1}.
\end{equation}
However, $\ell_1$ loss function cannot fully measure the modeling capacity from the view of video coding performance. It regards each pixel as an independent one and thus cannot measure the bits needed for coding the prediction residue, which is the other important factor that affects the final coding performance.

\begin{table}[t]
	\scriptsize
	\caption{Summarization of modules and the corresponding constraints.}
	\label{tab:hierarchical_constraints}
	\begin{tabular}{lll}
		\hline
		Module                                                                             & Constrained Domain                                                                    & Explanation                                                                                    \\
		\hline
		\hline
		\begin{tabular}[c]{@{}l@{}}Bi-directional motion\\ feature extraction\end{tabular} & Spatial                                                                               & Multi-scale encoder decoder                                                                 \\
		\hline
		\begin{tabular}[c]{@{}l@{}}Multi-scale frame\\ interpolation\end{tabular}          & \begin{tabular}[c]{@{}l@{}}Spatial, quality, frequency\end{tabular} & \begin{tabular}[c]{@{}l@{}}It integrates all parts \\ to get results\end{tabular}   \\
		\hline
		\begin{tabular}[c]{@{}l@{}}Factorized kernel\\ estimation\end{tabular}             & Spatial                                                                               & As shown in Fig. 1 (e)                                                                          \\
		\hline
		\begin{tabular}[c]{@{}l@{}}Quality attentive\\ maps estimation\end{tabular}             & Quality                                                                               & \begin{tabular}[c]{@{}l@{}}Make the network aware \\ of the quality differences\end{tabular}         \\
		\hline
		SATD loss                                                                          & Spatial, frequency                                                                 & \begin{tabular}[c]{@{}l@{}}The signal difference after \\ a frequency transformation\end{tabular} \\
		\hline
	\end{tabular}
\end{table}

\begin{figure}[t]
	\centering
	\subfigure[$\ell_1=16$, $\ell_{S}=16$ ]{
		\includegraphics[width=70mm]{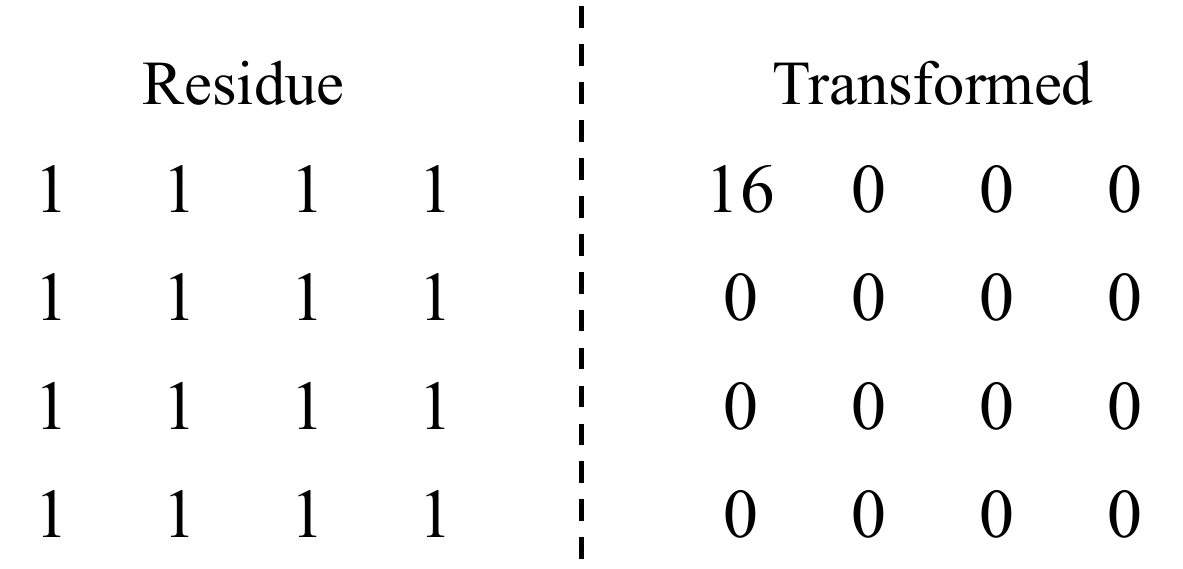}}
	\subfigure[$\ell_1=16$, $\ell_{S}=48$ ]{
		\includegraphics[width=70mm]{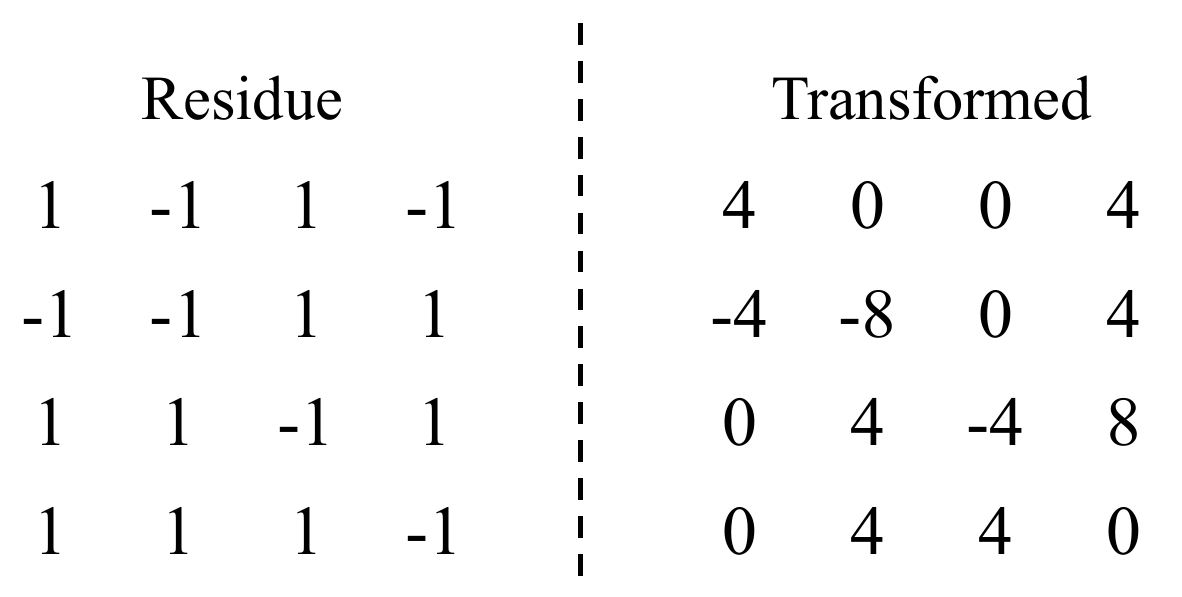}}
	\caption{Two example residue blocks with same $\ell_1$ losses but different $\ell_{S}$ losses. Intuitively, SATD loss is superior in measuring the redundancy of the residual signal after transform.}
	\label{fig:f3}
\end{figure}

In the fractional motion estimation process of HEVC, SATD is adopted as a matching criterion for it can better indicate the requirement of bits for coding residual signals. It is empirically proven that the numerical value of SATD after frequency transformation is more consistent with the number of bits to be spent for residual signals coding. Two example residue blocks and the corresponding $4 \times 4$ Hadamard transformed blocks are shown in Fig. \ref{fig:f3}. Though their $\ell_1$ losses are the same, the residue block (a) will intuitively cost less in the successive coding process since there are higher spatial similarities among the residual signals in the block. Compared with $\ell_1$, SATD successfully reflects the difference of the coding cost.

Consequently, we adopt SATD as the loss function $\ell_{S}$ to apply constraints to MSQ-FKCNN in the frequency domain for better coding performance. In conjunction with the hierarchical prediction architecture, multi-scale SATD loss function is further calculated to constrain the prediction process from coarse to fine in the frequency domain.

We calculate the $\ell_{S}$ loss by $8 \times 8$ blocks. The $8 \times 8$ Hadamard transformation matrix $\bf{H}$ is defined as follows:
\begin{equation}\label{eq4}
{\bf{H}} = \left[ {\begin{array}{*{20}{c}}
	{{\rm{1,}}}&{{\rm{1,}}}&{{\rm{1,}}}&{{\rm{1,}}}&{{\rm{1,}}}&{{\rm{1,}}}&{{\rm{1,}}}&{{\rm{1.}}}\\
	{{\rm{1,}}}&{{\rm{ - 1,}}}&{{\rm{1,}}}&{{\rm{ - 1,}}}&{{\rm{1,}}}&{{\rm{ - 1,}}}&{{\rm{1,}}}&{{\rm{ - 1.}}}\\
	{{\rm{1,}}}&{{\rm{1,}}}&{{\rm{ - 1,}}}&{{\rm{ - 1,}}}&{{\rm{1,}}}&{{\rm{1,}}}&{{\rm{ - 1,}}}&{{\rm{ - 1.}}}\\
	{{\rm{1,}}}&{{\rm{ - 1,}}}&{{\rm{ - 1,}}}&{{\rm{1,}}}&{{\rm{1,}}}&{{\rm{ - 1,}}}&{{\rm{ - 1,}}}&{{\rm{1.}}}\\
	{{\rm{1,}}}&{{\rm{1,}}}&{{\rm{1,}}}&{{\rm{1,}}}&{{\rm{ - 1,}}}&{{\rm{ - 1,}}}&{{\rm{ - 1,}}}&{{\rm{ - 1.}}}\\
	{{\rm{1,}}}&{{\rm{ - 1,}}}&{{\rm{1,}}}&{{\rm{ - 1,}}}&{{\rm{ - 1,}}}&{{\rm{1,}}}&{{\rm{ - 1,}}}&{{\rm{1.}}}\\
	{{\rm{1,}}}&{{\rm{1,}}}&{{\rm{ - 1,}}}&{{\rm{ - 1,}}}&{{\rm{ - 1,}}}&{{\rm{ - 1,}}}&{{\rm{1,}}}&{{\rm{1.}}}\\
	{{\rm{1,}}}&{{\rm{ - 1,}}}&{{\rm{ - 1,}}}&{{\rm{1,}}}&{{\rm{ - 1,}}}&{{\rm{1,}}}&{{\rm{1,}}}&{{\rm{ - 1.}}}
	\end{array}} \right]
\end{equation}
By dividing the residue $I_t-I_{m}$ into $T$ non-overlapping $8 \times 8$ residue blocks, we transform each residue block ${\bf{B}}_j$ by:
\begin{equation}\label{eq5}
{\bf{\tilde B}}_j = {\bf{H}} \times {\bf{B}}_j \times {\bf{H}},
\end{equation}
where ${\bf{\tilde B}}_j$ is the transformed residue block. Then, $\ell_{S}( {I_m,I_t} )$ can be obtained by sum of the absolute values of all the transformed residual signals:
\begin{equation}\label{eq6}
{\ell_{S}}\left( I_m,I_{t} \right) =
%\frac{1}{{T \times 8 \times 8}}
\sum\limits_{j = 1}^T {\sum\limits_{x = 1}^8 {\sum\limits_{y = 1}^8 {\left| {{{{\bf{\tilde B}}}_j}\left( {x,y} \right)} \right|} } }.
\end{equation}
The final multi-scale loss $\mathcal L$ is calculated by:
\begin{equation}\label{eq7}
{\cal L} = \alpha {\ell _{S}}( {I_m^{1/4},I_{t}^{1/4}} ) + \beta {\ell _{S}}( {I_m^{1/2},I_{t}^{1/2}} ) + \gamma {\ell _{S}}( {{I_m},{I_{t}}} ),
\end{equation}
where $\alpha, \beta, \gamma$ are the weighting parameters which are empirically set to $0.2,0.3,0.5$. The down-scaled images $I_{t}^{1/4}$ and $I_{t}^{1/2}$ are derived from $I_{t}$ with Bilinear interpolation.

\section{Training and Integration Details of MSQ-FKCNN}
\label{sec4}

\subsection{Training Data Preparation}
As for training data preparation, we use video clips to form the training samples. Each video clip consists of three consecutive frames $I_l$, $I_t$ and $I_r$, where $I_l$ and $I_r$ are the two-sided reference frames and $I_t$ is the ground truth. In the video coding scenario, two-sided reference frames are reconstructed frames which suffer from coding artifacts. The frame quality may be low especially for high QPs. In order to make the network work well in this condition, we code the reference frames $I_l$ and $I_r$ and use the reconstructed frames ${\hat I}_l$ and ${\hat I}_r$ as the input in training data generation. The reference frames are coded with HM-16.15 under the all intra configuration with a random QP value ranging from $0$ to $51$.

Besides, frames are coded under different QPs in RA configuration, which means two-sided reference frames usually have different quality. For the sake of further simulating the real application situation, we set QPs of two-sided reference frames to have a random difference of 0 to 10. QPs of the reference frames are also saved in the training set as the side information for training. With the quality attentive mechanism, our MSQ-FKCNN can be more aware of the quality difference between reference frames and learn to interpolate higher quality frames.

Later on, we randomly extract blocks with a size of $150 \times 150$ pixels at the same positions from two-sided coded reference frames and the ground truth frame to form the training data. A deep learning-based optical flow estimation method SpyNet\cite{spynet} is utilized here for candidate blocks selection. We will not add blocks whose mean optical flow values are large to the training set.

In the training process, we refer to \cite{Niklaus_ICCV_2017} for training data augmentation. $128 \times 128$ blocks are randomly cropped from the $150 \times 150$ blocks for training. The cropped blocks are also augmented by randomly changing the order of two-sided reference blocks and flipping all the blocks horizontally or vertically.

\begin{figure}[t]
\includegraphics[width=90mm]{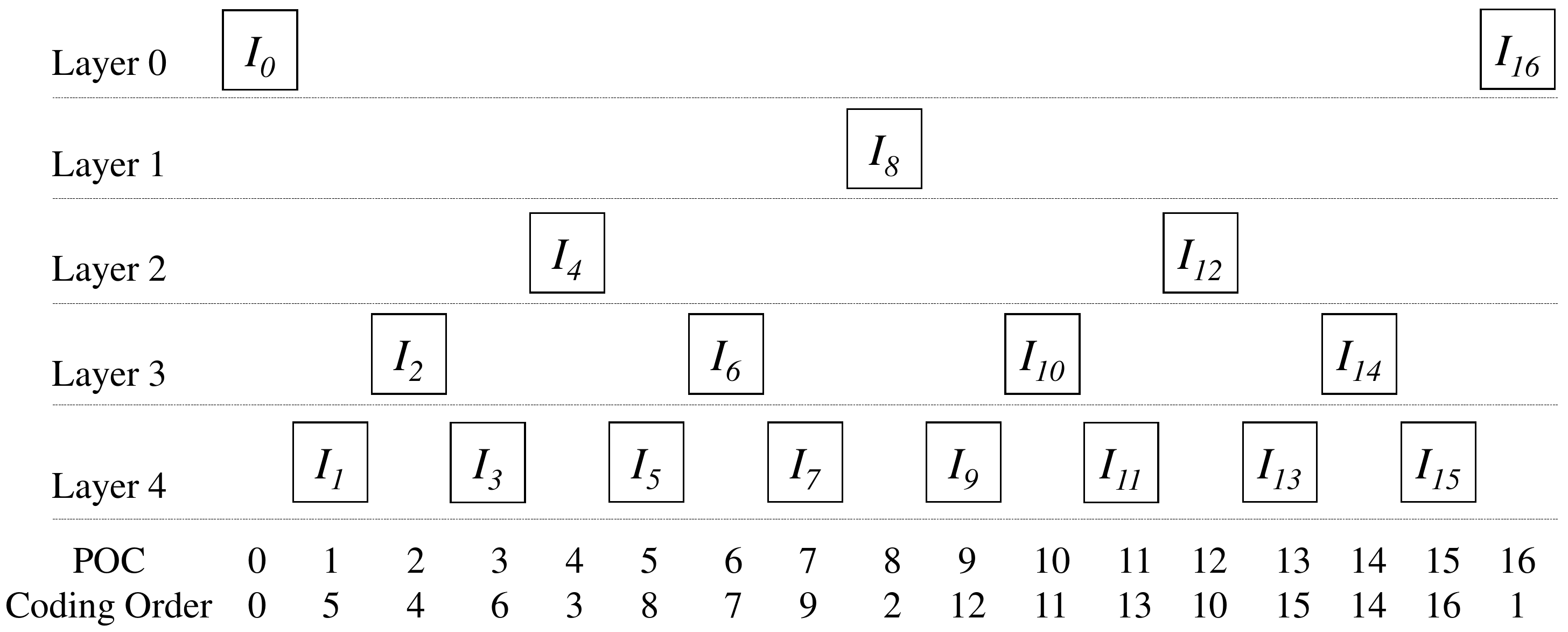}
\caption{Illustration of the hierarchical B coding structure in HM-16.15.}
\label{fig:f1}
\end{figure}

\subsection{Integration into HEVC}
We implement and test our method on HM-16.15 under the RA configuration, where frames are coded in the hierarchical B coding structure. Frames are allocated to different group of pictures (GOP) and frames of different GOPs are coded successively. In HM-16.15, each GOP consists of 16 frames. The coding order of frames in the same GOP is not decided by their picture order count (POC) value but systematically redesigned. As shown in Fig. \ref{fig:f1}, frames are assigned to different temporal layers. The frames are coded successively according to their temporal layers. Frames in higher layers can utilize the reconstructed frames in lower layers for inter prediction. Moreover, in addition to frames of the same GOP, coded frames in previous GOPs can also be adopted as the reference.

We choose to generate the PC-frame for frames whose temporal layers are greater than 1 in this paper. Specifically, for a to-be-coded frame $I_t$, we denote its temporal layer as $\tau \left( {{I_t}} \right)$ and the PC-frame can be generated as follows:
\begin{equation}\label{eq1}
{I_m} = \left\{ {\begin{array}{*{20}{c}}
{f\left( {{{\hat I}_{t - 4}},{{\hat I}_{t + 4}}} \right),\tau \left( {{I_t}} \right) = 2,}\\
{f\left( {{{\hat I}_{t - 2}},{{\hat I}_{t + 2}}} \right),\tau \left( {{I_t}} \right) = 3,}\\
{f\left( {{{\hat I}_{t - 1}},{{\hat I}_{t + 1}}} \right),\tau \left( {{I_t}} \right) = 4,}
\end{array}} \right.
\end{equation}
where $I_m$ is the desired PC-frame and ${\hat I}_{t + \ast}$ means the reconstructed reference frame. $f(\cdot)$ represents MSQ-FKCNN which infers the PC-frame from two-sided coded reference frames.

In the coding process, two reference picture lists $List0$ and $List1$ will be maintained. For most frames, two forward frames in $List0$ and two backward frames in $List1$ are available as reference for inter prediction.
For each reference frame, the reference frame index will be allocated to it which indicates its place in the reference picture list. Prediction units (PU) at the decoder side can find corresponding reference frames through decoded reference frame indexes. To add the interpolated PC-frame to reference picture lists, we choose an existing reference frame ${\hat I}_f$ in reference lists which is farthest from the to-be-coded frame $I_t$ and use its reference index to access $I_m$ at the decoder side.

${\hat I}_f$ and $I_m$ share the reference index of ${\hat I}_f$ in inter prediction. Specifically, we implement a CU level RDO to decide which reference frame to be accessed by the shared reference index.
Two passes of encoding that respectively use ${\hat I}_f$ and $I_m$ for inter prediction are performed at the decoder side.
A flag is set based on the rate-distortion costs of the two passes to indicate which reference frame to be used. When the flag is set to true, $I_m$ will be accessed if the shared reference index is chosen. Otherwise ${\hat I}_f$ will be used. The flag is coded with one bit and integrated at CU level. All PUs in a CU share the same flag. Moreover, if all PUs in a CU do not choose the shared reference index after two passes of encoding, we will not code the flag since it is no need to indicate which frame the shared reference index points to if it is never visited.

\begin{table}[t]
  \centering
    \caption{BD-rate reduction of the proposed method compared to HEVC.}

 % Table generated by Excel2LaTeX from sheet 'overall'
 \begin{tabular}{c|l|ccc}
 \hline
 \multirow{2}[4]{*}{Class} & \multicolumn{1}{c|}{\multirow{2}[4]{*}{Sequence}} & \multicolumn{3}{c}{BD-rate} \bigstrut\\
\cline{3-5}       &       & Y     & U     & V \bigstrut\\
 \hline
 \hline
 \multirow{3}[4]{*}{Class A} & Traffic & -6.1\% & -6.0\% & -4.6\% \bigstrut[t]\\
       & PeopleOnStreet & -11.0\% & -14.6\% & -12.6\% \bigstrut[b]\\
\cline{2-5}       & Average & -8.6\% & -10.3\% & -8.6\% \bigstrut\\
 \hline
 \hline
 \multirow{6}[4]{*}{Class B} & Kimono & -3.8\% & -5.7\% & -3.7\% \bigstrut[t]\\
       & BQTerrace & -0.6\% & -0.8\% & 0.1\% \\
       & BasketballDrive & -2.5\% & -4.5\% & -3.5\% \\
       & ParkScene & -5.1\% & -5.5\% & -4.1\% \\
       & Cactus & -5.4\% & -8.4\% & -7.4\% \bigstrut[b]\\
\cline{2-5}       & Average & -3.5\% & -5.0\% & -3.7\% \bigstrut\\
 \hline
 \hline
 \multirow{5}[4]{*}{Class C} & BasketballDrill & -5.2\% & -10.3\% & -9.9\% \bigstrut[t]\\
       & BQMall & -10.7\% & -13.9\% & -12.9\% \\
       & PartyScene & -7.4\% & -11.8\% & -9.6\% \\
       & RaceHorsesC & -2.4\% & -4.7\% & -4.7\% \bigstrut[b]\\
\cline{2-5}       & Average & -6.4\% & -10.2\% & -9.3\% \bigstrut\\
 \hline
 \hline
 \multirow{5}[4]{*}{Class D} & BasketballPass & -8.8\% & -11.0\% & -13.5\% \bigstrut[t]\\
       & BlowingBubbles & -6.5\% & -8.0\% & -7.6\% \\
       & BQSquare & -10.5\% & -6.4\% & -9.1\% \\
       & RaceHorses & -5.5\% & -8.2\% & -7.8\% \bigstrut[b]\\
\cline{2-5}       & Average & -7.8\% & -8.4\% & -9.5\% \bigstrut\\
 \hline
 \hline
 All Sequences & Overall & -6.1\% & -8.0\% & -7.4\% \bigstrut\\
 \hline
 \end{tabular}%
\vspace{-3mm}
\label{tab1}
\end{table}

\section{Experimental Results}
\label{sec5}

\subsection{Experimental Settings}
We use the Vimeo-90K dataset \cite{vimeo90k} to generate training data. The dataset consists of $89,800$ video clips with a fixed resolution of $448 \times 256$ resized from high-quality video frames. In total, $234,192$ samples are generated from the dataset for training. The network is implemented on PyTorch and AdaMax \cite{adam} is used as the optimizer with ${\beta _1} = 0.9,{\beta _2} = 0.999$. The learning rate is initially set to 0.001 and changed to 0.0001 after 30 epochs. We end the training process when 70 epochs are reached. The batch size is set to 16. We train our network on the Titan X GPU.

The proposed method is tested in HM-16.15 under the RA configuration with the intra period set to -1. BD-rate is used to measure the coding performance. HEVC common test sequences are adopted for testing. The number of encoding frames is set to be twice of the frame rate.
Four QP values $27$, $32$, $37$ and $42$ are employed in the experiment. It should be noted that we only need to train one model for all QPs. Luma and chroma components share the same interpolation model. During testing, the chroma components will be first up-sampled and concatenated with the luma component to form a three-channel YUV image. The YUV image is then transformed to an RGB image to form the input. We also compare with a method proposed in \cite{fruc}, which also introduces deep frame interpolation to video coding but directly use the interpolated block as the reconstruction block. For simplicity, we call it DVRF.

\begin{figure*}[t]
	\centering
\subfigure[PeopleOnStreet]{
	\includegraphics[width=0.45\linewidth]{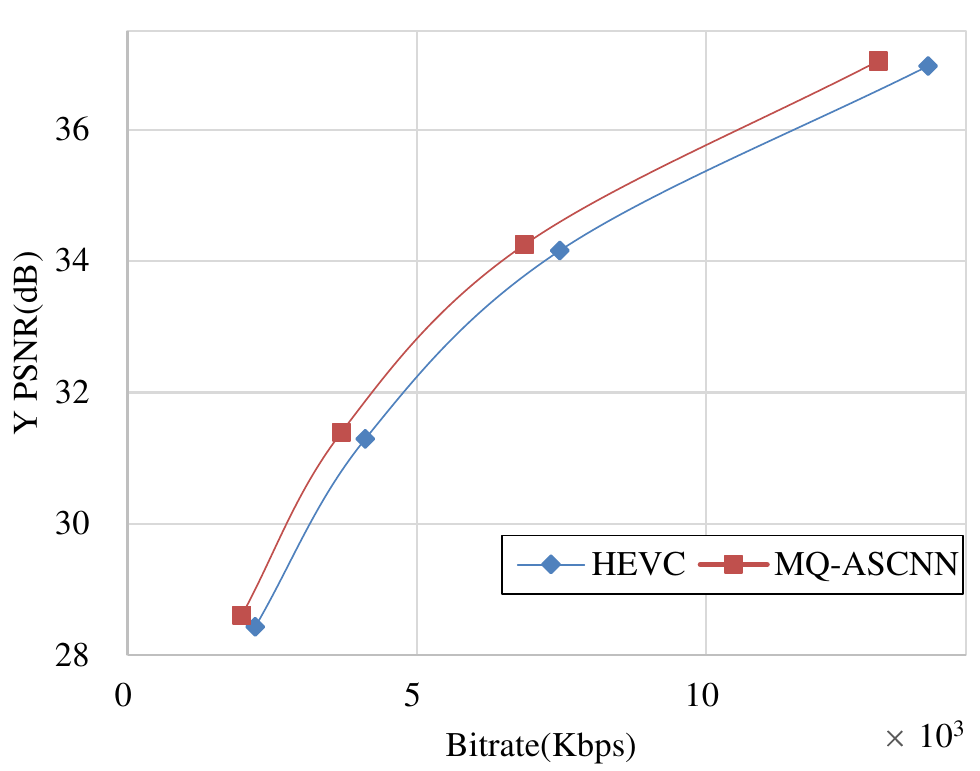}}
\subfigure[BQMall]{
    \includegraphics[width=0.45\linewidth]{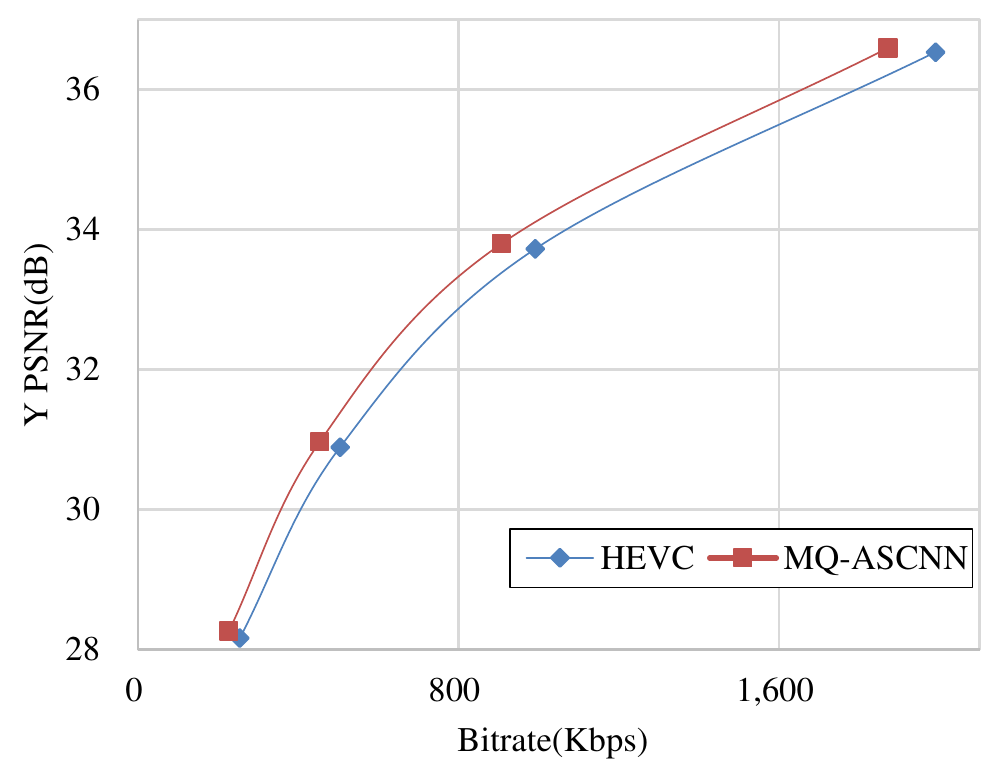}}
\subfigure[BasketballPass]{
	\includegraphics[width=0.45\linewidth]{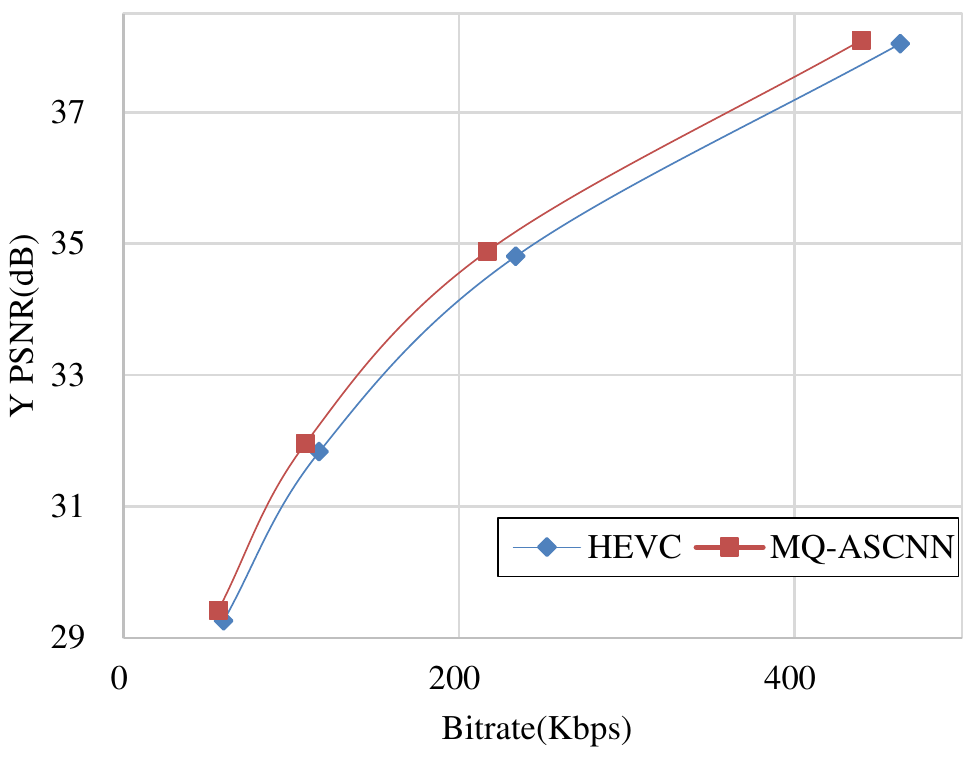}}
\subfigure[BQSquare]{
    \includegraphics[width=0.45\linewidth]{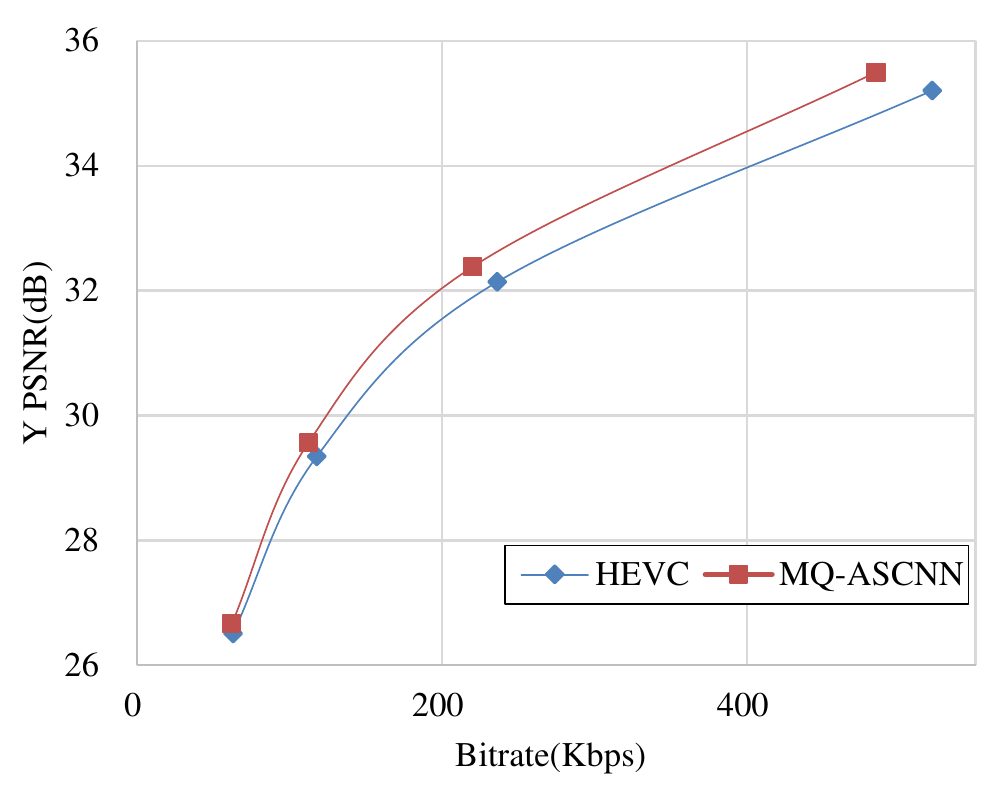}}
	\caption{Four example R-D curves of the sequences \emph{PeopleOnStreet}, \emph{BQMall}, \emph{BasketballPass} and \emph{BQSquare} for the luma component under RA configuration.}
	\label{fig:rdcurve}

\end{figure*}

\subsection{Experimental Results and Analysis}
\subsubsection{Overall Performance}
Table \ref{tab1} shows the overall performance of our method for classes A, B, C and D. Our method has obtained on average $6.1\%$, $8.0\%$ and $7.4\%$ BD-rate savings respectively for the Y, U, V components. For the test sequence \emph{PeopleOnStreet}, up to $11.0\%$ BD-rate saving can be obtained for the luma component. For further verification, some example rate-distortion (R-D) curves are shown in Fig.~\ref{fig:rdcurve}. It can be seen that our method is superior to HEVC under most QPs.

\begin{table}[htb]
 \centering

    \caption{BD-rate reduction comparison between DVRF and MSQ-FKCNN.}
  \begin{tabular}{c|l|c|c}
 \hline
 Class & \multicolumn{1}{c|}{Sequence} & DVRF  & Ours \bigstrut\\
 \hline
 \hline
 \multirow{6}[4]{*}{Class B} & Kimono & -1.7\% & -4.7\% \bigstrut[t]\\
       & BQTerrace & -0.2\% & -0.3\% \\
       & BasketballDrive & -1.1\% & -2.7\% \\
       & ParkScene & -2.6\% & -5.3\% \\
       & Cactus & -4.6\% & -6.1\% \bigstrut[b]\\
\cline{2-4}       & Average & -2.0\% & -3.8\% \bigstrut\\
 \hline
 \hline
 \multirow{5}[4]{*}{Class C} & BasketballDrill & -3.2\% & -5.6\% \bigstrut[t]\\
       & BQMall & -6.0\% & -10.1\% \\
       & PartyScene & -3.0\% & -6.3\% \\
       & RaceHorsesC & -0.8\% & -2.0\% \bigstrut[b]\\
\cline{2-4}       & Average & -3.2\% & -6.0\% \bigstrut\\
 \hline
 \hline
 \multirow{5}[2]{*}{Class D} & BasketballPass & -5.4\% & -9.9\% \bigstrut[t]\\
       & BlowingBubbles & -4.1\% & -6.0\% \\
       & BQSquare & -7.1\% & -9.0\% \\
       & RaceHorses & -2.2\% & -6.0\% \\
       & Average & -4.7\% & -7.7\% \bigstrut[b]\\
 \hline
 \hline
 All Sequences & Overall & -3.2\% & -5.7\% \bigstrut\\
 \hline
 \end{tabular}%
\vspace{-3mm}
\label{tab2}
\end{table}

\begin{figure*}[htbp]
	\centering
\subfigure[Target frame: Racehorses POC 1, left reference QP: 29, right reference QP: 40]{
	\includegraphics[width=0.4\linewidth]{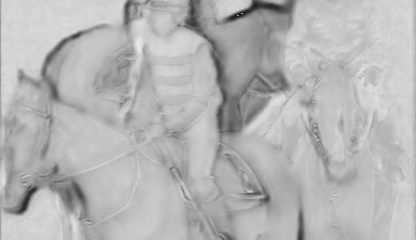}
	\includegraphics[width=0.4\linewidth]{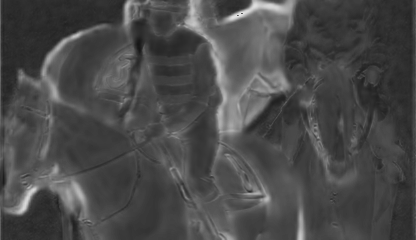}}
\subfigure[Target frame: Racehorses POC 13, left reference QP: 38, right reference QP: 40]{
    \includegraphics[width=0.4\linewidth]{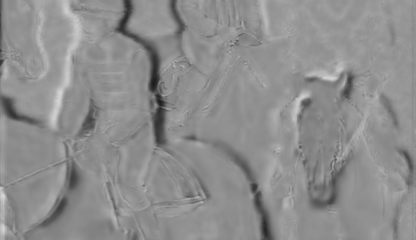}
    \includegraphics[width=0.4\linewidth]{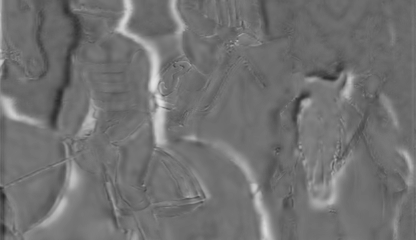}}
	\caption{Visualization examples of the weighting maps which indicate the proportion different reference frames take in the target frame interpolation. Brighter pixels mean higher weightings.}
	\label{fig:QAv}

\end{figure*}

\subsubsection{Comparison with the Existing Method}
Furthermore, we compare our MSQ-FKCNN with DVRF \cite{fruc}, which introduces a deep frame interpolation method to video coding. DVRF is implemented on HM-16.6. For a fair comparison, we also implement our method on HM-16.6 and test our method under the same conditions as DVRF. In the RA configuration of HM-16.6, the GOP size is 8 and the frames are divided into four temporal layers. Following DVRF, we also only deal with frames of layer 2 and layer 3 and directly replace the temporally farthest reference frame without CU level RDO.
% you can edit first.
%It should be noted that our method will achieve better performance if we also generate PC-frames for frames of layer 1 and integrate the PC-frames by CU level RDO.

%DVRF directly uses a model trained for frame interpolation to generate intermediate blocks and adopt them as the reconstructed blocks.

As shown in Table \ref{tab2}, though DVRF obtains gain over HEVC, they use a pre-trained model without any consideration on the video coding scenario, whose performance is limited. Moreover, directly utilizing generated blocks as the reconstructed blocks cannot fully exploit the benefits of frame interpolation and will bring prediction errors to the following coding process. Differently, by specially designing our model in the video coding scenario and integrating the generated PC-frame into inter prediction, our method obtains on average $2.5\%$ more BD-rate saving for the luma component compared with DVRF.

\subsubsection{Verification of Multi-Domain Hierarchical Constraints}
The effectiveness of multi-domain hierarchical constraints is also verified. A network named FKCNN is first implemented without the quality attentive mechanism and multi-scale frame interpolation. Q-FKCNN is later trained by adding the quality attentive mechanism to FKCNN to verify the quality attentive mechanism. Both FKCNN and Q-FKCNN are trained with the same settings as MSQ-FKCNN. The effectiveness of the hierarchical constraints can be proven by comparing between Q-FKCNN and MSQ-FKCNN.

\begin{table}[h]
	\scriptsize	
    \centering
    \caption{BD-rate reduction comparison for the verification of multi-domain hierarchical constraints.}
 \begin{tabular}{c|l|c|c|c}
 \hline
 Class & \multicolumn{1}{c|}{Sequence} & FKCNN & Q-FKCNNN & MSQ-FKCNN \bigstrut\\
 \hline
 \hline
 \multirow{5}[4]{*}{Class C} & BasketballDrill & -3.3\% & -3.4\% & -3.6\% \bigstrut[t]\\
       & BQMall & -8.4\% & -8.7\% & -9.2\% \\
       & PartyScene & -5.1\% & -5.1\% & -5.3\% \\
       & RaceHorsesC & -1.8\% & -2.1\% & -2.2\% \bigstrut[b]\\
\cline{2-5}       & Average & -4.7\% & -4.8\% & -5.1\% \bigstrut\\
 \hline
 \hline
 \multirow{5}[2]{*}{Class D} & BasketballPass & -6.0\% & -6.7\% & -8.5\% \bigstrut[t]\\
       & BlowingBubbles & -5.3\% & -5.6\% & -6.6\% \\
       & BQSquare & -6.8\% & -5.4\% & -7.4\% \\
       & RaceHorses & -3.9\% & -4.6\% & -5.0\% \\
       & Average & -5.5\% & -5.6\% & -6.9\% \bigstrut[b]\\
 \hline
 \hline
 All Sequences & Overall & -5.1\% & -5.2\% & -6.0\% \bigstrut\\
 \hline
 \end{tabular}%

\label{tab3}
\end{table}

Note that in the following comparison, we test all the sequences with 32 frames and only the first frame is coded under the all intra configuration. The comparison results of different networks on classes C and D for the luma component are shown in Table~\ref{tab3}.

It can be seen that a considerable BD-rate reduction can be obtained by adding the quality attentive mechanism to FKCNN for most test sequences. We additionally visualize the fusion weighting maps of the two-sided synthesized results for further verification of our quality attentive mechanism. The weighting maps are generated by dividing the synthesized results from left and right reference frames with the interpolated frame, which indicate the proportion each reference frame takes in the final result. The visualization results are shown in Fig. \ref{fig:QAv}. As we can see, reference frames of higher quality usually take a larger proportion in the final results. Moreover, the greater the quality difference is, the more proportion the higher quality one will obtain.

By comparing between Q-FKCNN and MSQ-FKCNN, we can find that on average 0.8\% and up to 2.0\% BD-rate reduction can be brought by employing the hierarchical constraints, which brings more multi-domain dependencies for reference and leads to more accurate prediction results.

\subsubsection{Verification of SATD Loss Function}
The superiority of $\ell_{S}$ loss function is also proven by experiments. We denote the models trained with $\ell_1$ and $\ell_{S}$ losses as MSQ-FKCNN-$\ell_1$ and MSQ-FKCNN-$\ell_{S}$, respectively. Table \ref{tab4} shows the BD-rate reduction obtained by models trained with different loss functions. By additionally constraining the interpolation in frequency domain, we can obtain on average $0.8\%$ more BD-rate reduction for the luma component for sequences of classes C and D.

\begin{figure*}[t]
	\centering
\subfigure[BasketballPass]{
	\includegraphics[width=0.23\linewidth]{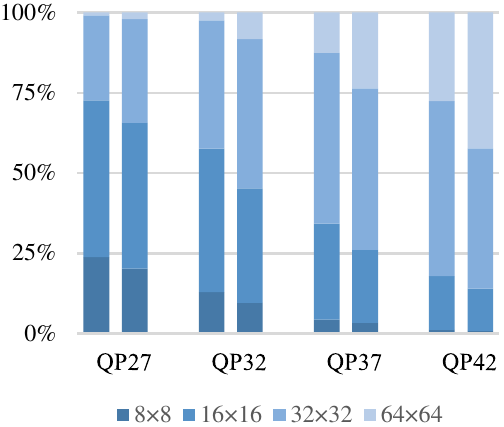}}
	\subfigure[BlowingBubbles]{
	\includegraphics[width=0.23\linewidth]{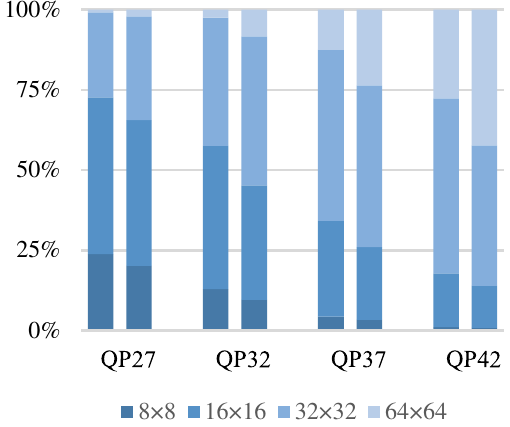}}
	\subfigure[BQSquare]{
    \includegraphics[width=0.23\linewidth]{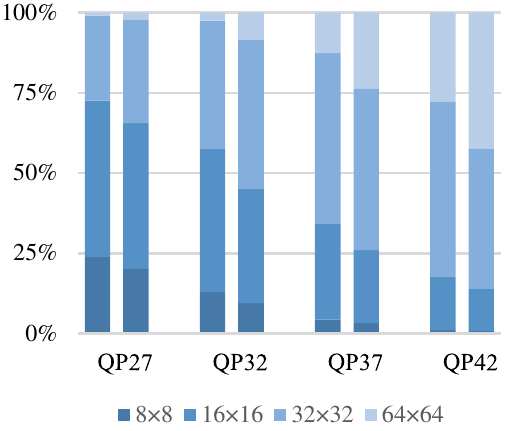}}
    \subfigure[RaceHorses]{
    \includegraphics[width=0.23\linewidth]{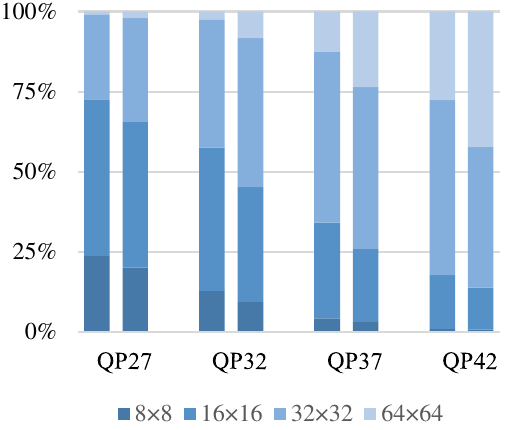}}
	\caption{Changes of the CU partition before and after using generated PC-frames for inter prediction. In each set, the left one shows ratios of different types of pixels coded by HM and the right one shows ratios of the pixels coded by our proposed method.}
	\label{fig:cusize}
	
\end{figure*}

\begin{table}[tbp]

  \centering
   \small
    \caption{BD-rate reduction comparison between models trained with different loss functions.}
 \begin{tabular}{c|ccc|ccc}
 \hline
 \multirow{2}[4]{*}{Class} & \multicolumn{3}{c|}{MSQ-FKCNN-$\ell_1$} & \multicolumn{3}{c}{MSQ-FKCNN-$\ell_{S}$} \bigstrut\\
\cline{2-7}       & Y     & U     & V     & Y     & U     & V \bigstrut\\
 \hline
 \hline
 Class C & -4.6\% & -8.0\% & -7.8\% & -5.1\% & -8.0\% & -7.6\% \bigstrut\\
 \hline
 Class D & -5.7\% & -7.2\% & -8.2\% &  -6.9\% & -7.3\% & -7.7\% \bigstrut\\
 \hline
 All   & -5.2\% & -7.6\% & -8.0\% & -6.0\% & -7.6\% & -7.6\% \bigstrut\\
 \hline
 \end{tabular}%
\label{tab4}

\end{table}

\begin{table}[t]

  \centering
   \small
    \caption{Ratios of CUs that choose PC-frames for inter prediction under different QPs.}
 \begin{tabular}{l|c|c|c|c|c}
 \hline
 \multirow{2}[4]{*}{Sequence} & \multicolumn{4}{c|}{Choosing Ratio} & \multirow{2}[4]{*}{BD-rate} \bigstrut\\
\cline{2-5}       & 27    & 32    & 37    & 42    &  \bigstrut\\
 \hline
 \hline
 BasketballPass & 40.4\% & 48.0\% & 51.7\% & 52.7\% & -8.8\% \bigstrut\\
 BlowingBubbles & 51.6\% & 60.3\% & 61.8\% & 44.1\% & -6.5\% \bigstrut\\
 BQSquare & 64.5\% & 70.8\% & 71.9\% & 40.2\% & -10.5\% \bigstrut\\
 RaceHorses & 22.3\% & 26.9\% & 34.8\% & 40.9\% & -5.5\% \bigstrut\\
 \hline
 \end{tabular}%
\label{tab5}
\end{table}

\begin{table}[t]

  \centering
   \small
    \caption{BD-rate reduction under the LDB configuration.}
 \begin{tabular}{c|l|ccc}
 \hline
 \multirow{2}[4]{*}{Class} & \multicolumn{1}{c|}{\multirow{2}[4]{*}{Sequence}} & \multicolumn{3}{c}{BD-rate} \bigstrut\\
\cline{3-5}       &       & Y     & U     & V \bigstrut\\
 \hline
 \hline
 \multirow{5}[4]{*}{Class C} & BasketballDrill & -2.1\% & -9.4\% & -6.9\% \bigstrut[t]\\
       & BQMall & -6.5\% & -11.1\% & -10.7\% \\
       & PartyScene & -3.3\% & -9.9\% & -7.8\% \\
       & RaceHorsesC & -0.6\% & -1.1\% & -0.8\% \bigstrut[b]\\
\cline{2-5}       & Average & -3.1\% & -7.9\% & -6.6\% \bigstrut\\
 \hline
 \hline
 \multirow{5}[4]{*}{Class D} & BasketballPass & -4.0\% & -8.5\% & -6.7\% \bigstrut[t]\\
       & BlowingBubbles & -3.1\% & -7.2\% & -9.4\% \\
       & BQSquare & -3.3\% & -6.6\% & -3.4\% \\
       & RaceHorses & -0.6\% & -1.2\% & -1.3\% \bigstrut[b]\\
\cline{2-5}       & Average & -2.7\% & -5.9\% & -5.2\% \bigstrut\\
 \hline
 \hline
 \multirow{4}[4]{*}{Class E} & FourPeople & -6.9\% & -8.5\% & -5.9\% \bigstrut[t]\\
       & Johnny & -3.7\% & -3.0\% & -2.6\% \\
       & KristenAndSara & -4.9\% & -7.3\% & -5.0\% \bigstrut[b]\\
\cline{2-5}       & Average & -5.2\% & -6.3\% & -4.5\% \bigstrut\\
 \hline
 \hline
 All Sequences & Overall & -2.9\% & -6.9\% & -5.9\% \bigstrut\\
 \hline
 \end{tabular}%
\label{tab6}
\end{table}

% \subsubsection{Different Frame Intervals}

\subsubsection{Rate Distortion Optimization and CU Partition Results Analysis}
For further verification of the proposed method, we analyze the RDO  and CU partition results on the sequences of class D. It should be noted that only frames whose temporal layers are greater than 1 are covered in our analysis. We first calculate the ratio of the CUs that choose PC-frames for inter prediction. The ratios are shown in Table \ref{tab5}. It can be seen that PC-frames generated by our MSQ-FKCNN are adopted by a considerable number of CUs for inter prediction.

Intuitively, more larger CUs will be used if we successfully alleviate the inconsistent pixel-wise displacement, since it is no need to further divide the CUs to handle the local differences caused by pixel-wise displacement. So we further analyze changes of the CU partition results before and after using the generated PC-frames. We divide pixels into four types according to sizes of the CUs they belong to. Later, ratios of different types of pixels are calculated and shown in Fig.~\ref{fig:cusize}. It can be found that more larger CUs have been used for inter prediction after adding PC-frames to the reference lists.

\subsubsection{Results under LD Configuration}
Furthermore, to test the generality of the proposed method, we also test our method under the low delay (LDB) configuration. We additionally train MSQ-FKCNN for the LDB configuration on newly prepared training data. Video clips containing three consecutive frames are used to form the training samples. In each clip, the first two frames are used to form the input and the third frame is used as the target. Testing results on sequences of classes C, D and E are shown in Table \ref{tab6}. On average 2.9\% BD-rate reduction can be obtained for the luma component under the LDB configuration.

\section{Conclusion}
\label{sec6}
In this paper, we propose a deep learning  based  frame  interpolation method to improve the inter prediction performance of HEVC. We carefully analyze the difficulties of frame interpolation encountered in the video coding scenario and pertinently propose the MSQ-FKCNN based frame interpolation regularized by multi-domain hierarchical constraints. The multi-scale quality attentive factorized kernel convolution is implemented to interpolate the target frame from small to large with quality attention. For the training of MSQ-FKCNN, multi-scale SATD loss function is employed to guide the network optimization in both spatial and frequency domains, which further improves the coding performance. After adding the generated PC-frames under the hierarchical B coding structure, significant BD-rate reduction can be obtained. Extensive experiments identify the effectiveness of each component in our MSQ-FKCNN and demonstrate the superiority of MSQ-FKCNN to the previous method.

\bibliographystyle{IEEEtran}
\bibliography{refs}

\end{document}